\documentclass[12pt,journal,compsoc,onecolumn]{IEEEtran}

\usepackage{graphicx}
\usepackage{comment}
\usepackage{amsmath,amssymb} 
\usepackage{color}
\usepackage{times}
\usepackage{epsfig}
 \usepackage{adjustbox}
 \usepackage{slashbox}
\usepackage{nicefrac}     
\usepackage{microtype}      
\usepackage{graphicx}
\usepackage{multirow}
\usepackage[ruled,lined,linesnumbered,boxed]{algorithm2e}

\usepackage{algorithmic}
\usepackage[noadjust]{cite}

\newtheorem{proposition}{Proposition}

\def\X{{\bf X}}

\def\U{{\bf U}}

\def\V{{\cal V}}

\def\F{{\cal F}}

\def\x{{\bf x}}

\usepackage{pifont}

\ifCLASSINFOpdf
\else
\fi

\begin{document}
 
\title{Frugal Reinforcement-based  Active Learning}

\author{\IEEEauthorblockN{Sebastien Deschamps$^{1,2}$ \ \ \ \ \ \ \ \ \  \ \ \ \ Hichem Sahbi$^{1}$ \\ $ $ \\}
\IEEEauthorblockA{$^{1}$ Sorbonne University, UPMC, CNRS, LIP6, F-75005 Paris, France \\ \ \ \ \ \ \ \ \    $^{2}$ Theresis Thales, France}}

\maketitle

\begin{abstract}
Most of the existing  learning models, particularly deep neural networks,  are reliant on
large datasets whose hand-labeling is expensive and time
demanding. A current trend is to make the learning of these
models frugal and less dependent on large collections of labeled data.
Among the existing solutions,  deep active learning is
currently witnessing a major interest and its purpose is to
train deep networks using as few labeled samples as
possible. However, the success of  active learning is
highly dependent on how critical are these samples when
training  models.  
In this paper, we devise a novel   active learning
approach for label-efficient training. The proposed method
is iterative and aims at minimizing a constrained objective
function that mixes diversity, representativity and
uncertainty criteria. The proposed approach is 
probabilistic and unifies all these criteria in a single
objective function whose solution models the probability of
relevance of samples (i.e., how critical) when
learning a decision function. We also introduce a novel
weighting mechanism based on reinforcement learning, which
adaptively balances these criteria at each training
iteration, using a particular stateless Q-learning model.
Extensive experiments conducted on staple image
classification data, including Object-DOTA, show the
effectiveness of our proposed model w.r.t. several
baselines including random, uncertainty and flat as well as other work.
\end{abstract}
 
\IEEEpeerreviewmaketitle

\section{Introduction}
 
Visual recognition aims at translating the  content of a given image into semantic categories \cite{Deng2009,Conf2008,theref2008,cvpr2008aaaa}.  This task is currently witnessing a tremendous interest in pattern recognition and image processing  through the use of deep learning models, and particularly  convolutional neural networks (CNNs) \cite{Krizhevsky2012,Szegedy2016,Szegedy2017} and more recently transformers \cite{Ashish2017}.  Nonetheless,  the success of these  models is  highly dependent on the availability of large collections of hand-labeled training  data.  In practice,   labeling manually large datasets is very time and effort demanding,   and the current trend is to {\it frugally} train models using transfer learning \cite{TL1}, domain adaptation \cite{DA1}, data augmentation \cite{DA2},  zero/few shot learning \cite{FSL,ref2015aaaa},   self-supervision \cite{SS1} and synthetic data/ground truth generation \cite{SGT1}.  However, the relative success of these solutions relies upon a strong assumption that knowledge are enough in order to close the {\it accuracy gap} while actually labeled data are  more important \cite{BMVC2012aaa,JSTARS2017aaa}. \\
 \indent Another category of methods is active learning \cite{Burr2009} which reserves the labeling effort  only to critical data, i.e.,  on well selected  and most qualitative subsets whose impact on the accuracy of the learned models is the most significant.   This process is iterative and asks an oracle (annotator) to label a few samples deemed informative from a large pool of unlabeled data,  prior to update a decision function that eventually maximizes generalization.   Most of  the active learning solutions are basically heuristics \cite{ICCV2007sahbi,M1,M2,M3,Sanjoy2004,Burr2009,GRSL2016sahbi}   which select unlabeled data by considering relevance measures that capture how critical are these data when learning decision functions. These measures are usually based on diversity,  representativity and uncertainty \cite{AL9}.   Diversity allows exploring different modes of data distribution while representativity seeks to select prototypical samples in each mode in order to  avoid outliers.  Uncertainty is instead used to locally refine the learned functions around ambiguous samples.  A suitable tradeoff between these criteria makes it possible to balance exploration and exploitation, two widely known concepts in active learning \cite{EXPEXP},  and this tradeoff is dependent on  the distribution of the data and the task at hand. \\ 
\indent In this paper, we introduce a novel active learning solution based on the minimization of a constrained objective function that mixes diversity, representativity and uncertainty criteria.  In contrast to most of the aforementioned existing solutions (and those described in section \ref{sec:related}), which are basically heuristics,  the proposed contribution is probabilistic and unifies all these criteria in a single objective function whose solution models the probability of relevance of samples (i.e., how critical are  samples) when learning  a decision function.  We also introduce a novel weighting process based on reinforcement learning (RL), which adaptively tradeoffs these criteria at each iteration of active learning,  and thereby avoids the combinatorial aspect of setting these criteria under the regime of frugal labeling.  The proposed RL approach relies on a particular stateless Q-learning model.  Extensive experiments conducted on  challenging image classification datasets,  including Object-DOTA, show the effectiveness of our proposed model w.r.t.  several related works including flat, random and uncertainty display model selection.

\section{Related work}
\label{sec:related}
\indent  Early active learning solutions  are based on Bayesian inference \cite{M1}, meta learning \cite{M2,Ref41,Ref37,Ref39} and more recent ones are dedicated to deep learning \cite{M3,Ref35,Ref36}. While some of these methods have shown a relative gain w.r.t. random sampling \cite{Ref13} and other baselines \cite{Ref14}, they are basically heuristics and lack groundedness.  In general,  state-of-the-art active learning algorithms include pool-based and generative models.  Pool-based methods use different acquisition strategies to sample informative examples among a pool of unlabeled data.  This category includes diversity \cite{Ref50a,Ref51a} and uncertainty-based techniques \cite{SS1,Ref54,Ref56,Ref57,Culotta2005} as well as their combination \cite{Ref47,Ref35,jordan2019}.   A representative work in diversity \cite{sener2017} casts the problem of active learning as a core-set selection \cite{Ref47a},  and proceeds by optimizing an euclidean distance between selected and non-selected data. The goal is to choose a subset of unlabeled data such that a model trained on it would perform similarly to a model trained on the whole dataset.   However,  core-set methods reach their limitation when distances between data become confound in high-dimensional spaces.  Other methods, based on uncertainty \cite{david1994,Ref40,Xin2013,Culotta2005},  attempt to select samples deemed ambiguous, with the assumption that the more uncertain a model is about its prediction, the more informative the sample is for that model.  For instance,  authors in \cite{david1994} select the most uncertain (and hence informative) samples, with the least confidence scores, using minimal margin and entropy on top of softmax class probabilities \cite{Ajay2009}.  However,  in spite of being widely used, softmax may not reflect the actual uncertainty in the learned models \cite{Ref53}.   Besides, using only uncertainty, particularly in batch-based CNNs, may result into redundant sampling which may lead to worse performances compared to random samples. Hence, several works attempt to combine uncertainty with diversity in order to overcome this limitation \cite{Ref47,Ref35,jordan2019}.  \\
\indent Another category of methods relies on generative adversarial networks (GANs) in order to synthesize informative training samples \cite{Ref31,Ref32}.  A variant known as cGAN  \cite{Ref48a}  conditions GAN on real images whereas ASAL \cite{Ref32} uses generated images to select/add similar real-world images to the training set together with their labels.  In the latter,  authors combine uncertainty, adversarial sample generation and matching to synthesize uncertain data. This is achieved, without an exhaustive search over pools of unlabeled data, with supposedly more resilience to sampling bias compared to other generative adversarial active learning approaches.  In GAAL \cite{Jia2017},  authors annotate synthetic samples and use them for linear SVMs and deep convolutional (DC) GANs training.  Nevertheless, their  method performs worse than random sampling; this is due to the sampling bias and also the difficulty in annotating the generated (poor quality) samples.  Overall,  the gain of these GAN-based  approaches has, thus far, not been consistently established w.r.t.  other strategies including random sampling \cite{Ref13}, maximal entropy and minimal distance baselines \cite{Ref40,Ref41,yoram2004},  and other approaches \cite{Ksenia2017,Sheng2010,Naoki1998} as well as self-taught learning \cite{Longlong2020}.
\def\X{\cal X}  
\def\L{\cal L}  
\def\U{\cal U}  
\def\D{{\cal D}}
 \def\Y{{\cal Y}}
 \def\x{{\bf  x}}
\def\C{{\bf C}}
\def\F{{\bf F}}
\def\DD{{\bf D}}
\def\tr{{\bf tr}}
\def\1{{\bf 1}}
\def\O{{\textrm{oracle}}}
 \def\diag{{\textrm{diag}}}

\section{Proposed model}
\label{sec:proposed}
Let $\X$ denote the set of all possible  images drawn from an existing but unknown probability distribution $P(X,Y)$.  In this definition,  the random variable $X$ refers to an input image and $Y$ to its unknown class label.  Considering $nc$ visual classes (a.k.a. labels or categories), and $\U$ as a large subset of $\X$ whose labels are initially unknown, our goal is to design classifiers $\{g_c\}_{c=1}^{nc}$ by interactively labeling  a very {\it small} fraction of $\U$, and  training the parameters of $\{g_c\}_c$.  This interactive labeling and training  is known as active learning.\\ 
\indent Let  $\D_t$ be a {\it display} (defined as a subset of $\U$) shown to an oracle\footnote{The oracle is defined as an expert annotator providing labels for any given subset of images.} at any iteration $t$ of active learning, and let $\Y_t$ be the underlying labels.   The initial display $\D_t$ (with $t=0$) is  uniformly sampled at random, and used to train  the subsequent classifiers  by repeating the following steps till reaching high generalization performances or exhausting a  labeling budget: 
\begin{itemize} 
\item Get the labels of  $\D_t$ as $\Y_t \leftarrow \O(\D_t)$.   This oracle function may depend on an {\it only-user-known}  ground-truth. 
\item Train $\{g_{c,t}\}_c$ using $\bigcup_{\tau=1}^t (\D_\tau,\Y_\tau)$, where the second subscript in  $g_{c,t}$ refers to the decision function at  iteration $t$.  In the remainder of this paper, different learning models will be considered including deep convolutional networks.  
\item Select the next display $\D \subset {\U}-\bigcup_{\tau=1}^t \D_\tau$ that possibly increases the generalization performances of the subsequent classifiers $\{g_{c,t+1}\}_c$. As the labels of the display $\D$ are unknown and also expensive,  one cannot combinatorially sample all the possible subsets  $\D$,  train the associated classifiers, and select the best display.  Alternative  selection strategies (a.k.a display models) are usually related to active learning and seek to find the most representative display that eventually yields optimal decision functions.  Nonetheless, one should be cautious in the way these sampling strategies are applied as many of them may lead to equivalent or worse performances compared to simple random sampling (see for e.g. \cite{reff2} and references within). 
\end{itemize} 

In what follows, we introduce our main contribution: a novel display  model which allows selecting the most representative samples to label by an oracle.   The proposed approach relies both on a constrained objective function and a weight selection strategy based on reinforcement learning.  This whole model turns out to be highly effective compared to different related display selection strategies including random,  uncertainty as well as  other related work   as corroborated  later in experiments. 
\subsection{Display selection model}
\indent We consider a probabilistic framework  which defines for each sample $\x_i \in \U$ a membership value $\mu_i$ that measures how likely is ``$\x_i$ belongs to subsequent  display $\D_{t+1}$''; consequently, $\D_{t+1}$ will correspond to the unlabeled data in $\{\x_i\}_i \subset \U$ with the highest memberships $\{\mu_i\}_i$. Considering $\mu \in \mathbb{R}^n$ (with $n= |{\U}|$) as a vector of these memberships  $\{\mu_i\}_i$, we propose to find $\mu $ as the optimum of the following constrained minimization problem 

{\begin{equation}\label{eq0}
  \begin{array}{l}
    \displaystyle    \min_{\mu \geq 0, \|\mu\|_1=1}  \eta \ \tr\big(\diag(\mu' [\C \odot \DD])\big) + \alpha  \ [\C' \mu]' \log [\C' \mu] \\
             \ \ \ \ \ \ \  \ \ \ \ \ \ \ \ \ \ \    \ \   + \beta \  \tr\big(\diag(\mu' [\F \odot \log \F]) \big) + \gamma \ \mu' \log \mu, 
 \end{array}  
\end{equation}}
\noindent here $\odot$, $'$ are respectively the Hadamard product and the matrix transpose, $\|.\|_1$ is the $\ell_1$ norm, $\log$ is applied entry-wise, and $\diag$ maps a vector to a diagonal matrix.  In the above objective function
\begin{itemize} 
\item $\DD \in \mathbb{R}^{n \times K}$ and $\DD_{ik}=d_{ik}^2$ is the euclidean distance between $\x_i$ and $k^{\textrm{th}}$ cluster centroid of a partition of $\U$ ($\{h_1,\dots,h_K\}$) obtained with K-means clustering. 
\item $\C \in \mathbb{R}^{n \times K}$ is a binary indicator matrix with each entry  $\C_{ik}=1$ iff $\x_i$ belongs to the $k^{\textrm{th}}$ cluster, and $0$  otherwise. 
\item  And $\F \in \mathbb{R}^{n \times nc}$ is a scoring matrix with $\F_{ic} =\hat{g}_{c,t}(\x_i)$ and $\{\hat{g}_{c,t}\}_c$  being a stochastic variant of the initial decision functions $\{{g}_{c,t}\}$, i.e.,  $\hat{g}_{c,t}(.)  \in [0,1]$ and   $\sum_c \hat{g}_{c,t} (.)=1$.  In the particular context of deep convolutional networks,  these normalized classifiers  correspond to  softmax  layer.  
\end{itemize}

The  first term $\tr\big(\diag(\mu' [\C \odot \DD])\big)$ in Eq. \ref{eq0} (equal to $\sum_i \sum_k  1_{\{\x_i \in h_k \}} \mu_i d_{ik}^2$) measures the {\it representativity} of the selected samples in $\D$; it captures how close is each data $\x_i$ w.r.t. the centroid of its cluster,  and this term reaches its smallest value when the centroids are sufficiently numerous and when they coincide with the selected samples.  The second  term $[\C' \mu]' \log [\C' \mu] $ in Eq. \ref{eq0}  (equivalent to  $\sum_{k}   [\sum_{i=1}^n 1_{\{\x_i \in h_k \}} \mu_i] \log [\sum_{i=1}^n 1_{\{\x_i \in h_k \}} \mu_i]$) captures the {\it diversity} of the selected samples, defined as the entropy of the probability distribution of the underlying clusters; this term is minimized when the selected samples belong to different clusters and vice-versa. The third criterion  $\tr\big(\diag(\mu' [\F \odot \log \F]) \big)$ (equal to $\sum_i \sum_c^{nc} \mu_i \F_{ic} \log \F_{ic} $) captures  the {\it ambiguity} in $\D$ measured by the entropy of $\{\hat{g}_{c,t}(.)\}_c$; this third term reaches it smallest value when  data are evenly scored w.r.t.  different categories. Finally, the fourth term is related to the {\it cardinality} of $\D$, measured by the entropy of the distribution $\mu$; without any a priori about the three other criteria, the fourth term favors a flat $\mu$-distribution and  acts as a regularizer. 

\subsection{Optimization}
\begin{proposition}
The optimality conditions of (\ref{eq0}) lead to the solution 
\begin{equation}\label{eq1}
  \mu^{(\tau+1)} :=\displaystyle \frac{\hat{\mu}^{(\tau+1)} }{\|\hat{\mu}^{(\tau+1)}\|_1}, \\
\end{equation} 
with $\hat{\mu}^{(\tau+1)}$ being  
\begin{equation}\label{eq2}
  \exp\bigg(-\frac{1}{\gamma}[\eta (\DD\odot \C)\1_K + \alpha \C (\log[\C' {\mu}^{(\tau)}]+\1_K)+\beta (\F \ \odot \  \log \F)\1_{nc}] \bigg), 
\end{equation}
here $\1_{nc}$, $\1_{K}$ denote two vectors of $nc$ and $K$ ones respectively. \\
\end{proposition} 
\def\CC{{\cal C}}

\noindent Details of the proof are omitted and  result from the gradient optimality conditions of Eq.~(\ref{eq0}).   Considering the above proposition, the optimal solution is obtained iteratively as a fixed point of Eqs (\ref{eq1}) and (\ref{eq2}) with $\hat{\mu}^{(0)}$ initially set to random values. Note that convergence is  observed in practice in few iterations, and the underlying fixed point, denoted as $\tilde{\mu}$, corresponds to the most {\it relevant} samples in the display $\D_{t+1}$ (according to criterion~\ref{eq0}) used to train the subsequent classifier  $\{\hat{g}_{c,t+1}\}_c$ (see also algorithm~\ref{alg1}).  The setting of $\gamma$ in Eq. \ref{eq2} controls the  sharpness  of the $\mu$-distribution; larger values result into flat distribution while smaller values to Dirac-like distribution.  A reasonable setting of $\gamma$ consists in dividing the numerator inside the exponential by its norm. \\
\begin{algorithm}[!ht]
\KwIn{Images in $\U$, display ${\cal D}_0 \subset {\U}$, budget $T$, $B$.}
\KwOut{$\cup_{t=0}^{T-1} (\D_t,\Y_t)$ and $\{g_t\}_{t}$.}
\BlankLine
\For{$t:=0$ {\bf to} $T-1$}{$\Y_t \leftarrow \textrm{oracle}(\D_t)$; \\ 
  $g_{t} \leftarrow \arg\min_{g} P(g(X)\neq Y)$ \tcp*[r]{Learning model (built on top of $\cup_{k=0}^t (\D_k,\Y_k)$)} 
 $\hat{\mu}^{(0)} \leftarrow \textrm{rand}$; $\mu^{(0)} \leftarrow \displaystyle \frac{\hat{\mu}^{(0)} }{\|\hat{\mu}^{(0)}\|_1}$; $\tau \leftarrow 0$ \\  
\BlankLine
 \While{($\|\mu^{(\tau+1)}-\mu^{(\tau)}\|_1\geq  \epsilon \ \wedge \ \tau<\textrm{maxiter})$}{
   Set ${\mu}^{(\tau+1)}$ using Eqs.~(\ref{eq1}) and (\ref{eq2}) \tcp*[r]{Display model} 
   $\tau \leftarrow \tau +1$
 }
$\tilde{\mu} \leftarrow \mu^{(\tau)}$ \\ 
$\D_{t+1} \leftarrow \{ \x_i \in {\U}\backslash \cup_{k=0}^t \D_k: \tilde{\mu}_i \in {\cal L}_B(\tilde{\mu})\}$ \tcp*[r]{${\cal L}_B(\tilde{\mu})$ being the $B$ largest values of $\tilde{\mu}$}
}
\caption{Display selection mechanism}\label{alg1}
\end{algorithm}
 
\noindent As shown  in the remainder of this paper, the setting of the other hyper-parameters $\alpha, \beta, \eta$  is crucial for the success of the  display model.  For instance, putting more emphasis on diversity (i.e.,  high $\alpha$) results into high exploration of class modes while a high  focus on ambiguity (i.e.,  large $\beta$) locally refines the trained decision functions.  A suitable balance between exploration  and local refinement of the learned decision functions should be achieved by selecting the best configuration of these hyper-parameters.  Besides, the setting of these hyper-parameters should be iteration-dependent as early,  intermediate and late iterations $t$ may require different  display selection strategies.  Moreover,  since labeling is sparingly achieved and on-the-fly,  no extra  labeled validation sets could be made  available {\it beforehand} in order to ``optimally'' set these hyper-parameters; and even when labeled validation sets are available,  tuning these hyper-parameters through all the iterations $t$  is highly combinatorial and intractable\footnote{This tuning is intractable  as the number of hyper-parameters scales linearly  w.r.t. the max number of iterations $T$, and the number of possible grid search configurations scales polynomially as ${\cal O}(p^T)$ where $p$ is the number of possible tested configurations for each hyper-parameter.}.  

\subsection{RL-based display selection}\label{rl}
In what follows, we rewrite the classifiers $\{{g}_{c,t}\}_c$ trained at a given iteration $t$ simply as ${g}_{t}$.  Let $\Lambda_\alpha$,  $\Lambda_\beta$, $\Lambda_\eta$ denote the parameter spaces associated to $\alpha, \beta, \eta$ respectively,  and let $\Lambda$  be the underlying   Cartesian product.    For any subsequent iteration $t+1$,  and for any instance  $\lambda_{t+1} \in \Lambda$ (written for short as $\lambda$),  one may obtain a display (now rewritten as  $\D^\lambda_{t+1}$) by solving   Eq.~\ref{eq0}, and the best configuration $\lambda^*$ that yields an optimal display could be defined as  
 \begin{equation}\label{eq33}
\lambda^* \leftarrow \arg\min_{\lambda \in \Lambda} {\cal R}_\textrm{emp}(g_{t+1};{\cal V}_t),
\end{equation} 
\noindent here ${\cal V}_t \subset \cup_{\tau=1}^t \D_\tau$ is a holdout set taken from the previous oracle's annotations\footnote{Note that classifiers $\{g_{t}\}_{t}$ are trained on $\{\cup_{\tau=1}^{t} \D_\tau\}_t$ but these training sets are deprived from $\{{\cal V}_t\}_t$, and the latter are used only for validation.} and $ {\cal R}_\textrm{emp}(g_{t+1};{\cal V}_t)$ denotes the empirical risk of $g_{t+1}$ on ${\cal V}_t$.   As solving  Eq.~\ref{eq33} requires  generating  and labeling multiple displays $\D^\lambda_{t+1}$  for different $\lambda$,  and training the underlying classifiers,  Eq.~\ref{eq33} makes finding the best configuration $\lambda^*$  clearly intractable.  Moreover,  in the frugal learning regime,  one may not afford labeling multiple displays; besides,  the holdout sets $\{{\cal V}_t\}_t$ are not sufficiently large in practice to make the setting of $\lambda^*$ reliable which may lead to weak generalization.  In order to bypass all these limitations, we consider in what follows an efficient and effective framework, based on RL, which allows training these hyper-parameters while considering not only the immediate reward (current classifier accuracy) but also future estimates of these rewards. \\

\noindent {\bf Hyper-parameter selection.} We consider an RL algorithm based on  Markov Decision Process (MDP) (see for instance  \cite{Sutton2018}).  The latter corresponds to a tuple $\langle S,A,R,q,\delta\rangle $ with $S$ being a state set, $A$ an action set,  $R: S \times A \mapsto \mathbb{R}$ an immediate reward function,  $q: S\times A \mapsto S$ a transition function and $\delta$ a discount factor.  An RL agent interacts with an environment by running a sequence of actions from $A$ with the goal of maximizing an expected discounted reward.   The agent follows a stochastic policy, $\pi: S \mapsto A$, which computes the true state-action value as  
\begin{equation}
Q(s,a) = E_\pi \left[  \sum_{t=0}^\infty  \delta^t {r}_t | S_0,=s,  A_0=a \right],
\end{equation} 
where ${r}_t=R(s,a)$ is an immediate reward at iteration $t$ of RL,  $S_0$ an initial state,  $A_0$ an initial action  and $\delta \in [0,1]$ is a discount factor that balances between immediate and future rewards. The goal of the optimal policy is to select actions that maximize the  discounted cumulative reward; i.e., $\pi_*(s) \leftarrow \arg\max_{a} Q(s,a)$ with $Q(s,\pi_*(s))$ being the optimal action value.  One of the most used methods to solve this type of RL problems is Q-learning \cite{Allen2018}, which directly estimates the optimal value function and obeys the fundamental identity, the Bellman equation 
\begin{equation}\label{bell}
 Q(s,a)  \leftarrow (1-\gamma_{rl})  Q(s,a) + \gamma_{rl}  (r_t + \delta  \max_{a'} Q(s',a')),
\end{equation} 
being  $\gamma_{rl}$  and $\delta$ the learning rate and the discount factor respectively set (in practice) to 0.1,   0.9 and $s'=q(s,a)$.  We consider in our hyper-parameter optimization, a stateless version,  so $Q(s,a)$,  $R(s,a)$ are rewritten simply as $Q(a)$ and $R(a)$ respectively.  
One may turn the optimization of the  hyper-parameters $(\alpha, \beta, \eta)$ either on continuous or  discrete domain  $\Lambda$.  In the continuous case,  $\Lambda$ is equal to $\mathbb{R}^3\backslash(0,0,0)$ and the underlying action set $A$ corresponds to  27 possible joint incremental updates of  $\alpha, \beta,\eta$  by three multiplicative factors   taken (in practice) from  $\{1,0.95, (0.95)^{-1}\}^3$.   In the discrete case, $\Lambda$ equates $\{0,1\}^3\backslash(0,0,0)$  so the underlying action set $A$ corresponds instead to 7 possible binary configurations of  $\alpha, \beta,\eta$.  At each iteration $t$,  the reward $R(a)$ of a given action $a \in A$ will be evaluated once the action executed and  the underlying subsequent display $\D_{t+1}$ and classifier $g_{t+1}$ trained.  Following Eq. ~\ref{eq33}, the reward $R(a)$ is measured using the accuracy  $ 1 - {\cal R}_\textrm{emp}(g_{t+1};{\cal V}_t)$ of the learned decision function $g_{t+1}$ evaluated on the holdout set $\V_{t} \subset \cup_{\tau=1}^t \D_\tau$ whose cardinality does not exceed (in practice) $10\%$ of the oracle's annotated displays. Note that this holdout set is used only for reward estimation (and hence hyper-parameter update) and not for classifier training.   The detailed steps of our  RL-based display selection are shown in algorithm~\ref{alg2}.

\begin{algorithm}[!ht]
\KwIn{Images in $\U$, display ${\cal D}_0 \subset {\U}$, budget $T$, $B$.}
\KwOut{$\cup_{t=0}^{T-1} (\D_t,\Y_t)$ and $\{g_t\}_{t}$.}
\BlankLine
$\forall \textrm{action}$,  $Q(\textrm{action}) \leftarrow \textrm{rand}$  \tcp*[r]{rand $\in [0,1]$} 
\For{$t:=0$ {\bf to} $T-1$}{
$\V_t \leftarrow    \textrm{RSubset}(\D_t)$  \tcp*[r]{Random subset} 
$\Y_t \leftarrow \textrm{oracle}(\D_t)$  \tcp*[r]{Oracle annotations}
  $g_{t} \leftarrow \arg\min_{g} P(g(X)\neq Y)$ \tcp*[r]{Learning model (built on top of $\cup_{k=0}^t \D_k\backslash \V_k$)}

          \If {$t\geq 1$}{
              $r_t \leftarrow 1-P(g_t(\V_t)\neq Y(\V_t))$ \tcp*[r]{Reward} 
  Set   $Q(\textrm{prev\_action})$ using the stateless version of Eq. \ref{bell}; \\
}
  $v \leftarrow \textrm{rand}$ ; \\
\If {$v\leq \exp(-t)$}{
            $\textrm{best\_action} \leftarrow \textrm{random\_action in} \ A$   \tcp*[r]{explore with prob $\exp(-t)$} 
        }
        \Else {      
  $\textrm{best\_action} \leftarrow \arg\max_\textrm{action} Q(\textrm{action})$ \tcp*[r]{otherwise take best Q-value} 
  }
  $\textrm{prev\_action} \leftarrow \textrm{best\_action}$ ; \\
  $(\alpha,\beta,\eta) \leftarrow \textrm{update(best\_action},\alpha,\beta,\eta)$ ; \\
 $\hat{\mu}^{(0)} \leftarrow \textrm{rand}$ ; $\mu^{(0)} \leftarrow \displaystyle \frac{\hat{\mu}^{(0)} }{\|\hat{\mu}^{(0)}\|_1}$ ; $\tau \leftarrow 0$ \\  
\BlankLine
 \While{($\|\mu^{(\tau+1)}-\mu^{(\tau)}\|_1\geq  \epsilon \ \wedge \ \tau<\textrm{maxiter})$}{
   Set ${\mu}^{(\tau+1)}$ using Eqs.~(\ref{eq1}) and (\ref{eq2}) \tcp*[r]{Display model} 
   $\tau \leftarrow \tau +1$
 }
$\tilde{\mu} \leftarrow \mu^{(\tau)}$ ; \\ 
$\D_{t+1} \leftarrow \{ \x_i \in {\U}\backslash \cup_{k=0}^t \D_k: \tilde{\mu}_i \in {\cal L}_B(\tilde{\mu})\}$ \tcp*[r]{${\cal L}_B(\tilde{\mu})$ being the $B$ largest values of $\tilde{\mu}$}
}
\caption{RL-based Display selection mechanism}\label{alg2}
\end{algorithm}
 
 \section{Experiments}
We study the impact of our proposed display selection model on two remote sensing tasks: satellite image change detection \cite{sahbiigarss2011,bourdis1,bourdis2} using the Jefferson dataset \cite{sahbi2021},  and remote sensing object classification using  Object-DOTA \cite{dota2018}.  In the first task, i.e.,  change detection, the goal is to find occurrences of targeted changes in satellite image pairs taken at different instants \cite{ICPR2016sahbi}. 
 The Jefferson dataset, used in change detection,  consists of 2,200  non-overlapping patch pairs (of 30 × 30 RGB pixels each). These pairs correspond to registered (bi-temporal) GeoEye-1 satellite images of 2,400 x 1,652  pixels with a spatial resolution of 1.65m/pixel, taken from the area of Jefferson (Alabama) in 2010 and in 2011. These images show multiple changes due to tornadoes in Jefferson (building destruction, etc.) as well as no-changes (including irrelevant ones as clouds). The underlying  ground-truth consists of 2,161 negative pairs (no/irrelevant changes) and only 39 positive pairs (relevant changes), so more than 98\% of this area corresponds to no-changes and this makes the task of finding relevant changes even more challenging. In our experiments, half of the dataset is used to train the display and the learning models while the remaining half for evaluation. \\
 \noindent The second dataset --- Object-DOTA  as a variant of DOTA \cite{dota2018} --- is larger  and used for image classification.  Object-DOTA contains 127,759 remote sensing  snapshots,  belonging to 15 categories (including harbors,  ships, etc.).  These snapshots were taken from 2,806 large remote sensing images, both in gray-scale and RGB, of dimensions ranging from 800$^2$ to 20,000$^2$ pixels.  DOTA images  were originally collected from   Google-Earth as well as   GF-2 and JL-1 satellites.  The number of images per category ranges from 98 to 37,028, so categories  are also highly imbalanced.   Training and test sets include 98,906  and 28,853 data respectively.    As classes in  both Jefferson and Object-DOTA are highly imbalanced, we  measure the classification performances using the equal error rate (EER); the latter is a balanced generalization error that evenly weights errors through different classes.  Smaller EER (or equivalently larger {\it accuracy} defined as 1-EER) implies better performances.
 \subsection{Backbones and pretraining} 
Images in Jefferson and Object-DOTA are encoded using two pretrained backbones; the GCN (graph convolutional network) in \cite{sahbi2021b} for the former,  and the ViT \cite{Dosovitskiy2020} for the latter.   The GCN consists of multiple blocks of aggregation and  inner product layers  followed by  pooling and fully connected layers.   Note that these networks are pretrained  differently; indeed,  the used GCN is pretrained on Jefferson, but using a self-supervised pretext loss similar to the one in \cite{Carl2015} while the ViT is pretrained on a different set (namely ImageNet \cite{Deng2009}) using a supervised loss. In both cases,  no ground-truth is used on Jefferson and Object-DOTA for pretraining.  
\subsection{Model Analysis and Ablation}
In order to study the impact of different terms of our objective function, we consider them individually, pairwise and all jointly taken. In this study, the last term of Eq. \ref{eq0} is always kept as it acts as a regularizer and allows obtaining the closed form in Eq. \ref{eq1}. The impact of each of these terms and their combination is shown in Tables \ref{tab1} and \ref{tab22}. From these results, we observe the highest impact of representativity+diversity especially at the earliest iterations of frugal learning, whilst the impact of ambiguity term raises later in order to locally refine the decision functions (i.e., once the modes of data distribution become well explored). These  performances are shown for different sampling percentages  at each  iteration $t$. 

\begin{figure}[h]
\begin{center}
  \begin{tabular}{cc}
 \includegraphics[width=.42\linewidth]{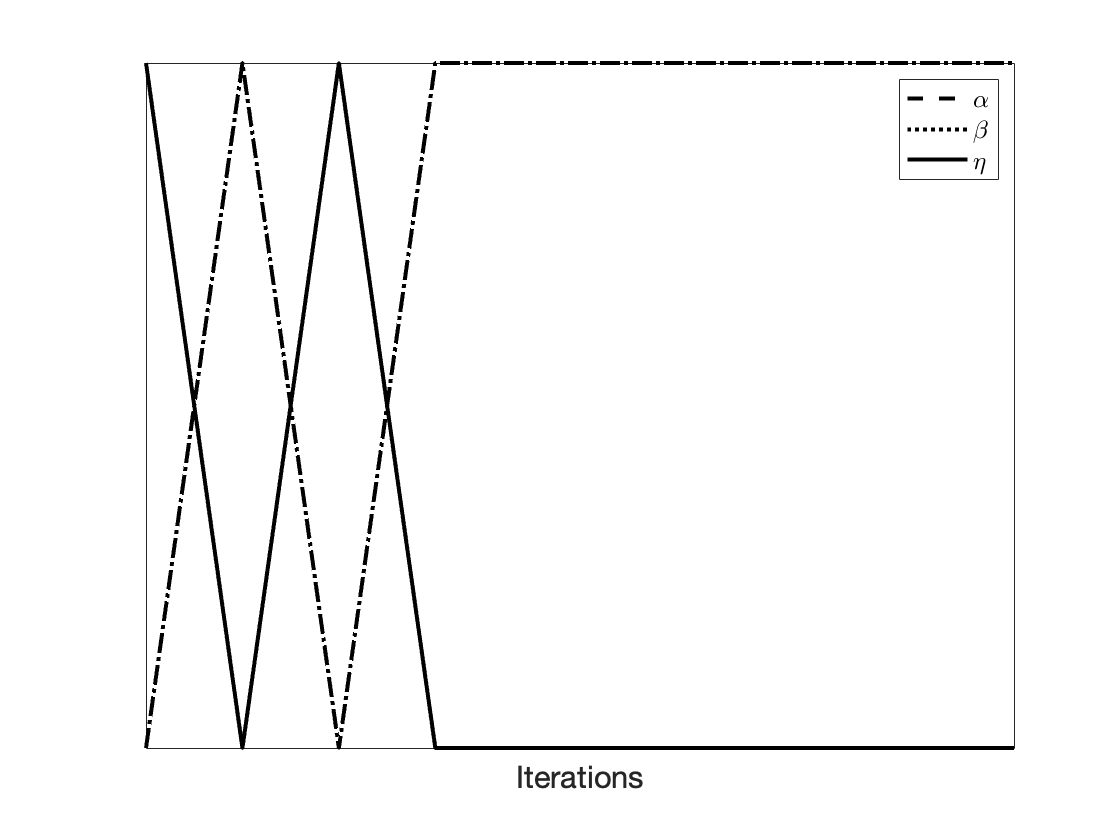} & \hspace{-1.02cm}  \includegraphics[width=.42\linewidth]{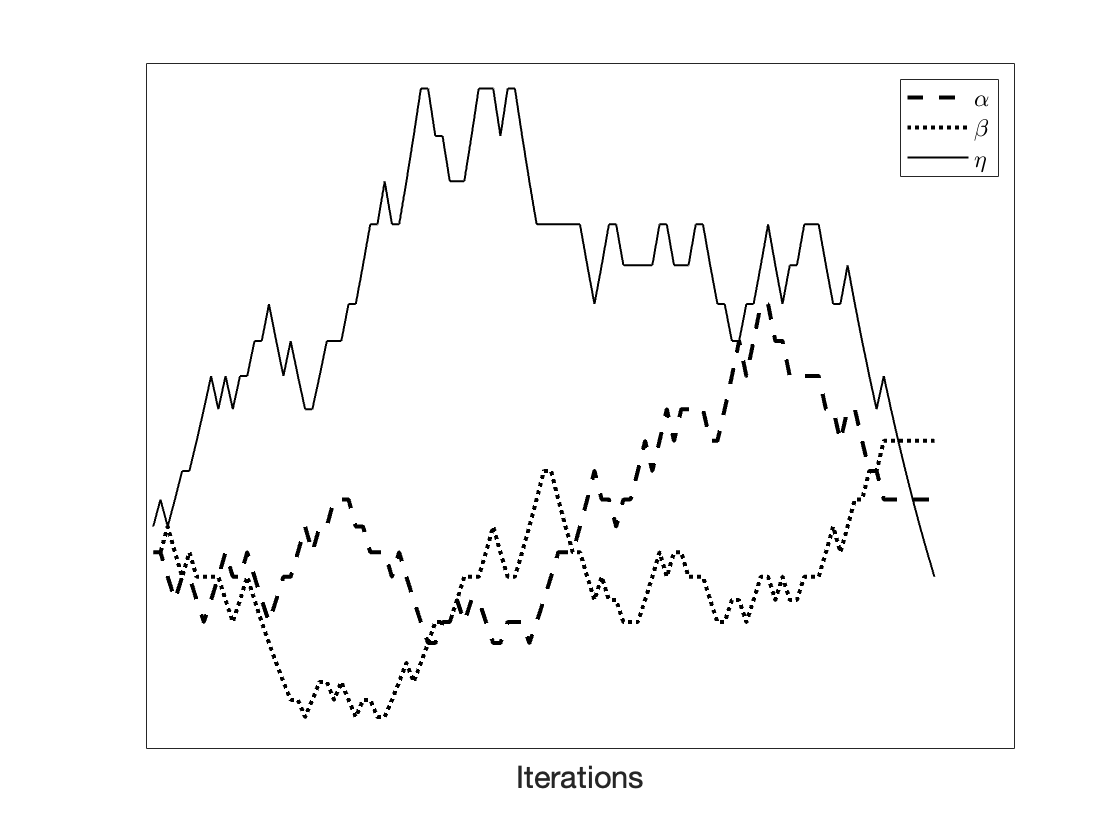} \\
 \includegraphics[width=.42\linewidth]{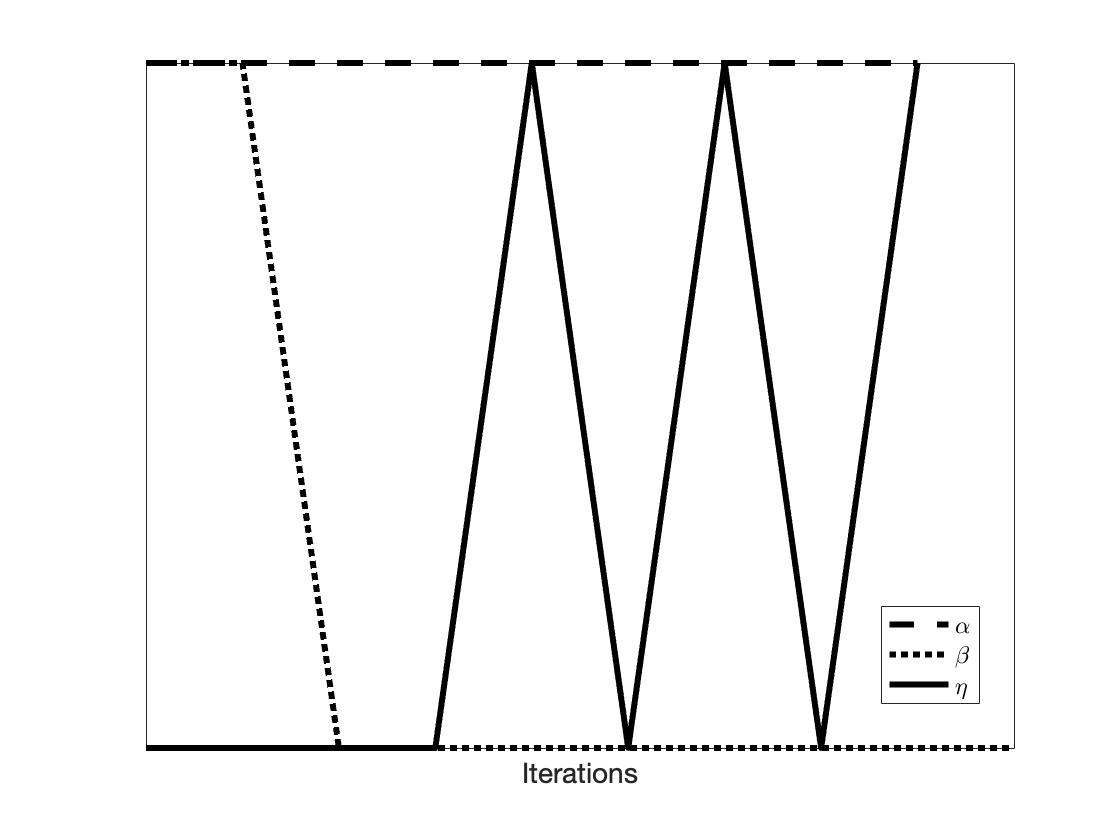} & \hspace{-1.02cm} \includegraphics[width=.42\linewidth]{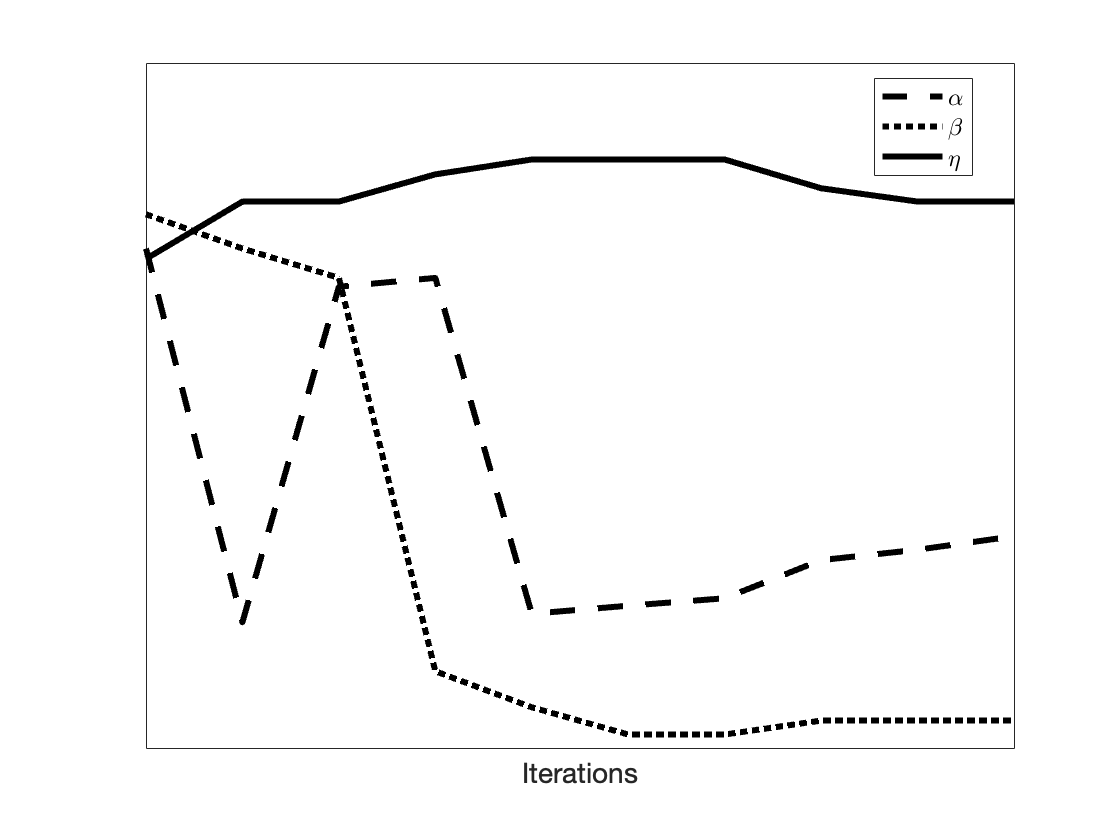}\\
  \end{tabular}
  \end{center}
 \caption{This figure shows the dynamic of the learned hyper-parameters w.r.t different iterations of active learning on Jefferson (top) and Object-DOTA (bottom). These  figures correspond to the discrete RL (left) and the continuous one (right).  As observed, the variation in the continuous setting is gradual while  in the discrete case the variation  is more abrupt.}\label{tab333}\end{figure}

According to these results, none of the settings (rows) in Table \ref{tab1} and Table \ref{tab22} obtains the best performance through all the iterations. Considering these observed ablation performances, a better setting of the  $\alpha,$  $\beta$ and $\eta$ should be iteration-dependent using  RL (as described in section \ref{rl}), and as also corroborated through performances shown Tables \ref{tab1}, \ref{tab22} and also the dynamics of the learned hyper-parameters through iterations $t$ (shown in Fig.~\ref{tab333}).  Indeed, it turns out that this adaptive setting outperforms the other combinations (including “all”, also referred to as “flat”), especially at the late iterations (highest sampling percentages) for RL-discrete  (RL-D), and  the low/mid sampling percentages for RL-continuous (RL-C) on Jefferson, and the late iterations on Object-DOTA.   Nevertheless, the average performance of RL-C is better than RL-D.

 \begin{table}[h!]
  \caption{This table shows an ablation study of our display model on Jefferson.  Here rep, amb and div stand for representativity, ambiguity and diversity respectively. These results are shown for different iterations $t$ (Iter) and the underlying sampling rates (Samp).  RL-D and RL-C refer to our proposed reinforcement based discrete and continuous models respectively.  {\bf In this table, smaller EERs imply better performances.}}\label{tab1} 
 \centering
 \resizebox{0.85\textwidth}{!}{
 \begin{tabular}{c||cccccccccc}
Iter  &2 & 3& 4& 5& 6& 7& 8& 9 & 10 \\  
Samp\%  &2.90 & 4.36& 5.81& 7.27& 8.72& 10.18& 11.63& 13.09 & 14.54  \\  
 \hline
 \hline
  rep &  26.21 &  12.72 &   10.48 &    9.88 &   9.70&    8.52 &   8.85&   8.61&   8.82 \\  
  div &  31.24 &  23.45 &  30.41 &  44.81   &  24.12 &   13.22 &   17.02&    6.88. &   7.98 \\  
 amb & 46.68 &   38.73 &   29.91 &   14.74 &   20.11 &   8.33 &   7.41 &   7.37 &   5.53  \\  
 rep + div &  26.21 &   33.35 &   25.10 &   21.55 &   11.71 &   2.84 &   1.65 &   1.59 &   1.43 \\  
 rep + amb &   26.21 &   12.62 &   10.81 &   9.82 &   9.70 &   8.53 &   9.23 &   8.60&   8.82 \\  
   div + amb &  41.69 &   28.82 &  23.08 &   23.41 &   23.42 &   19.82 &   13.10 &   8.16 &   6.97 \\  
   all (flat) &  26.21 &  33.35 &   25.52 &   23.70 &  14.59 &   2.74 &   1.54&   1.67   & 1.48 \\  
  \hline \hline
      RL-D  & 31.75&  \bf10.36&   14.83&   13.36&   14.70&   \bf1.06 & \bf1.06  & \bf1.10  &  \bf1.01  \\  
      RL-C & \bf9.91 &  21.29  &  \bf9.95 &   \bf6.54   & \bf4.65 &   2.63 &   2.44  & 2.95  &  1.80   
\end{tabular}}

 \end{table}

 \begin{figure}[h]
\begin{center}
  \includegraphics[width=0.75\linewidth]{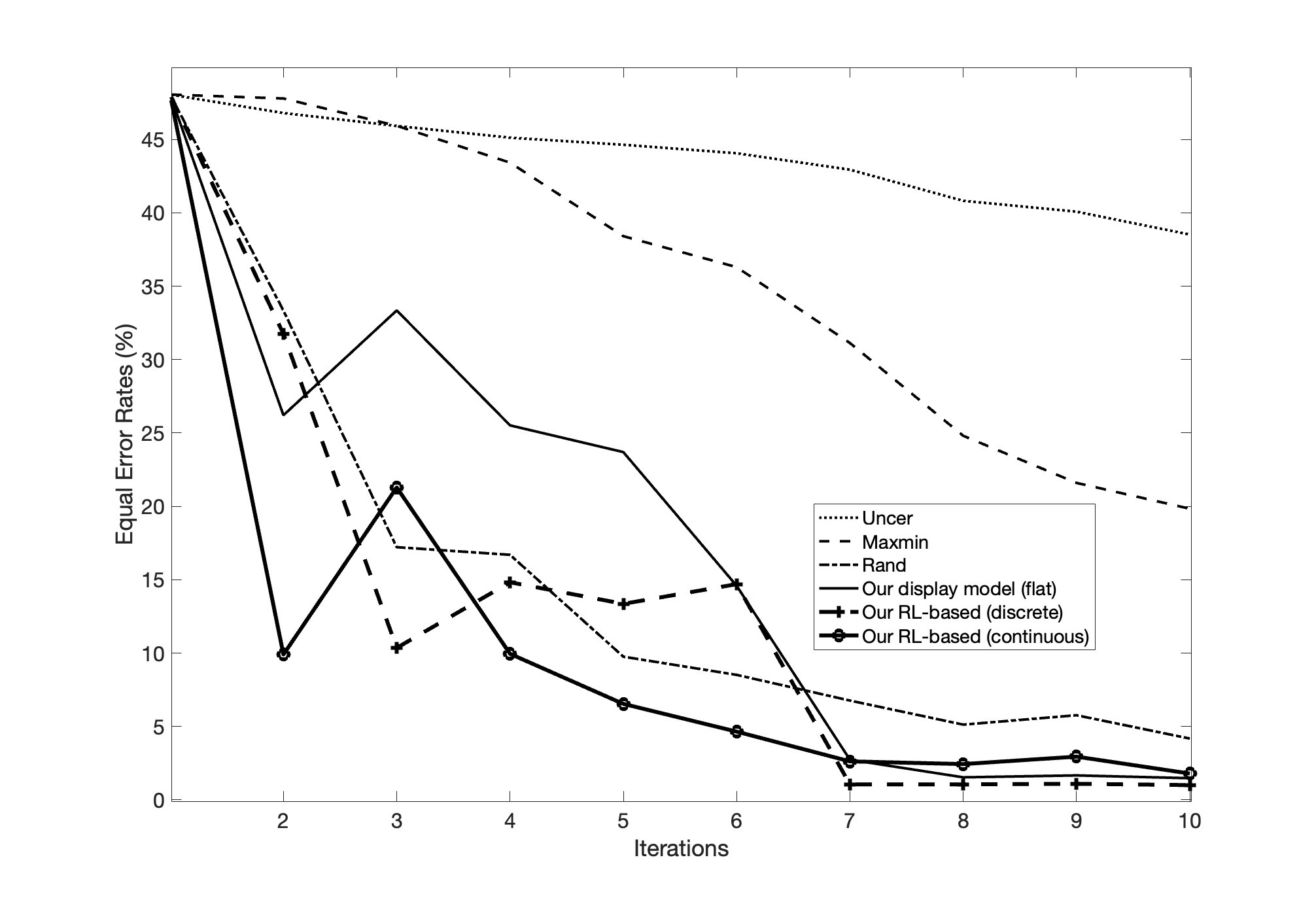}
\end{center}  
 \caption{This figure shows a comparison of different sampling strategies w.r.t. different iterations (Iter) and the underlying sampling rates in table~\ref{tab1} (Samp) on Jefferson. Here Uncer and Rand stand for uncertainty and random sampling respectively.  Note that fully-supervised learning achieves an EER of $0.94 \%$. See again section~\ref{comp} for more details.}\label{tab2}\end{figure}

 \begin{table}[h!]
  \caption{Same caption as Table.~\ref{tab1}, excepting experiments were achieved on Object-DOTA. Note that performances are reported via the accuracy (1-EER) with a higher frugal regime (i.e., lower percentages of labeled data compared to Jefferson data).  {\bf In this table, higher accuracies  imply better performances.}}\label{tab22} 
 \centering
 \resizebox{0.85\textwidth}{!}{
 \begin{tabular}{c||ccccccccc}
Iter  & 2 & 3& 4& 5& 6& 7& 8& 9 & 10  \\   
Samp\%  &  0.2  & 0.3& 0.4& 0.5& 0.6& 0.7& 0.8& 0.9 & 1.0 \\  
 \hline
 \hline
  rep &  25.77 &  27.91 &   29.81 &   30.69&  31.96&    33.51 &   34.08&   34.53&   35.05 \\  
   div &  26.16 &  34.04 &  37.97 &  41.33 &   41.85 &  44.44&    45.57&   48.12 & 48.72  \\  
 amb &  38.44 &   48.86  &   54.96 &   58.21 &   59.51 &   61.03 &   61.14 &   62.61 &   62.66 \\ 
  rep + div &  49.57 &   51.38 &   53.60 &   54.44 &   54.74 &   55.49 &   55.47 &  55.87 &   56.25 \\  
rep + amb &     42.04 &   49.43 &   53.49&   57.23& 59.73 &   61.88 &   62.78 &   63.42 &   64.16 \\  
    div + amb &   41.76 &   48.54 &  53.11 &   56.21 &   57.32 &  58.12 &   59.05 &   60.07 &   61.10 \\  
        all (flat)  &  47.18 &  56.80 &   59.73 &   61.03 &  63.70 &   63.85 &  64.34&  64.74   & 65.23 \\ 
  \hline \hline
               RL-D          & 35.42  & 41.19 & 43.44 & 46.28 & 51.29  & 53.43 & 54.09  & 53.47 &56.59 \\
          RL-C & 46.29&  55.36&   57.82&   60.72&  63.08&   \bf64.43 & \bf64.88  & \bf66.20  &  \bf66.61   
\end{tabular}}

 \end{table} 

 \begin{figure}[h]
\center
\includegraphics[width=0.75\linewidth]{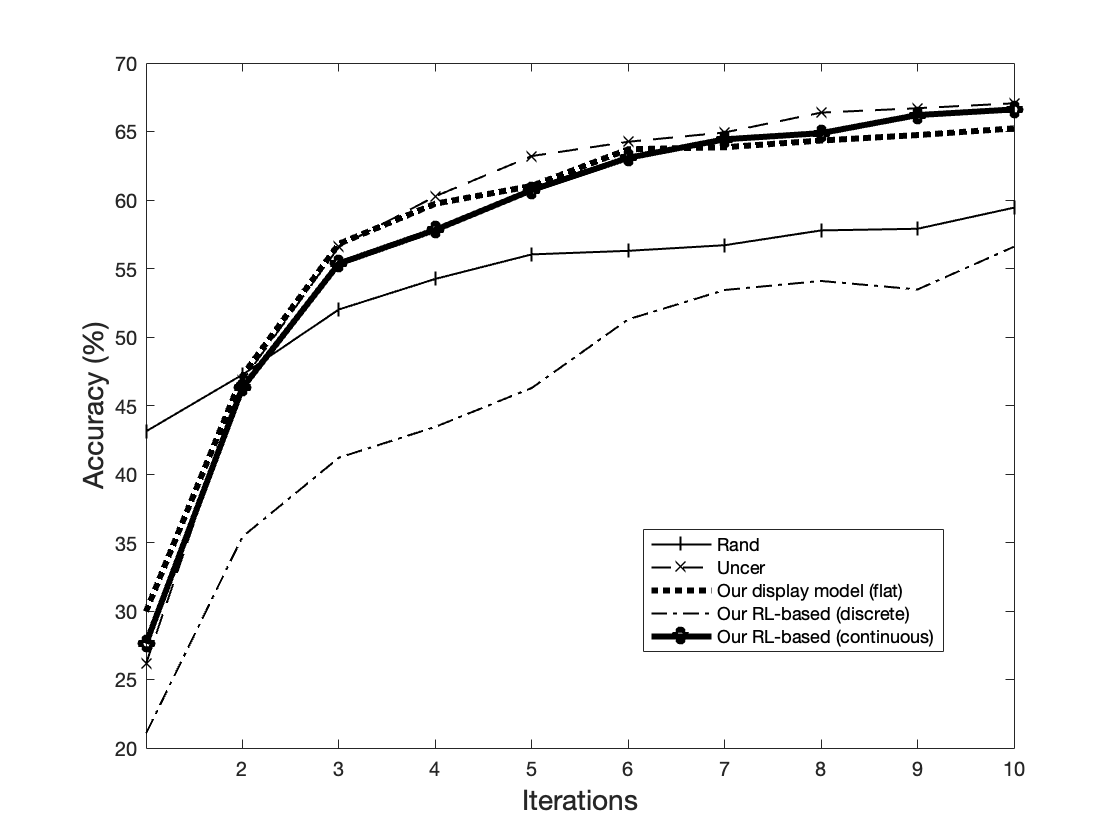}
 \caption{This figure shows a comparison of different sampling strategies w.r.t. different iterations (Iter) and the underlying sampling rates in table~\ref{tab22} (Samp) on Object-DOTA.  Note that fully-supervised learning achieves  an accuracy of $74.92 \%$.  See again section~\ref{comp} for more details.}\label{tab2222}\end{figure}
 
\subsection{Extra Analysis and Comparison}\label{comp}

Figure.~\ref{tab2} shows extra comparisons of our RL-based display model against different related display sampling strategies  including {\it random, MaxMin and uncertainty}.  Random selects data from ${\U}\backslash \cup_{k=0}^t {\cal D}_k$  whereas MaxMin (similar to \cite{sener2017}) greedily selects a sample $\x_i$  in ${\cal D}_{t+1}$ from the pool ${\U}\backslash \cup_{k=0}^t {\cal D}_k$ by maximizing its minimum distance w.r.t  $\cup_{k=0}^t {\cal D}_k$. We also compare our method w.r.t. uncertainty \cite{Culotta2005} which consists in selecting samples in the display whose scores are the most ambiguous (i.e.,  the closest to zero). Finally, we also consider the fully supervised setting as an upper bound on performances; this configuration relies on the whole annotated training set and builds the learning model in one shot. \\ 
The performances  in figure~\ref{tab2} (and also figure \ref{tab2222}) show the positive impact of the proposed RL-based display, both on the discrete and the continuous models,  against the related sampling strategies for different amounts of annotated data.  Excepting the flat model (also used in \cite{sahbi2021}),  most of these comparative methods are powerless to classify data sufficiently well.  Indeed, the comparative methods are effective either at the early iterations of active learning (such as MaxMin and random which capture the diversity of data without being able to refine decision functions) or at the latest iterations (such as uncertainty which locally refines decision functions but suffers from the lack of diversity).  The flat display strategy  \cite{sahbi2021} gathers the advantages of random, MaxMin and uncertainty, but suffers from the rigidity of the weights of  representativity,  diversity and ambiguity criteria which are fixed instead of being learned (i.e.,  iteration-dependent).  In contrast, our proposed RL-based design adapts the choice of these criteria as iterations evolve; it's worth noticing that RL-C is effective including at the early iterations, and this makes it  more suitable for high frugal regimes.   In sum,  the proposed RL-based display  makes classification   reaching lower EERs (and equivalently high accuracy) and overtakes all the other strategies at the end of the iterative  learning process.   

\section{Conclusion}
We introduce in this paper a novel display learning model based on the optimization of an objective function mixing representativity, diversity and ambiguity.  
The proposed approach is probabilistic and assigns membership measures to  unlabeled samples,  and selects the display as samples with the highest memberships. 
The proposed approach also relies on an RL-based mechanism which selects the best (discrete or continuous) combination of  representativity, diversity and ambiguity through active learning iterations, thereby leading to better performances. Extensive experiments conducted on the task of remote sensing image classification and change detection show the outperformance of the proposed method against different settings as well as  related work. \\
As a future work, we are currently investigating the use of self-supervised learning methods in order to further enhance the generalization capacity of our learning models as well as the use of generative networks for other display model design.

\end{document}